%% file: main.tex
\PassOptionsToPackage{table}{xcolor}
\documentclass[letterpaper]{article} 
\usepackage[draft]{aaai2026}  
\input{packages}

\title{ConfAgents: A Conformal-Guided Multi-Agent Framework for Cost-Efficient Medical Diagnosis}

\author{
    Huiya Zhao\textsuperscript{\rm 1, $\ast$},
    Yinghao Zhu\textsuperscript{\rm 1,2, $\ast$},
    Zixiang Wang\textsuperscript{\rm 1, $\ast$},
    Yasha Wang\textsuperscript{\rm 1},
    Junyi Gao\textsuperscript{\rm 3,4},
    Liantao Ma\textsuperscript{\rm 1, $\dagger$},
}
\affiliations{
    \textsuperscript{\rm 1}Peking University, \textsuperscript{\rm 2}The University of Hong Kong,\\
    \textsuperscript{\rm 3}The University of Edinburgh, \textsuperscript{\rm 4}Health Data Research UK\\
    yhzhu99@gmail.com, malt@pku.edu.cn
}

\begin{document}

\maketitle

\setcounter{footnote}{0}
\renewcommand{\thefootnote}{\fnsymbol{footnote}}
\footnotetext{$^\ast$ Equal contribution, $^\dagger$ Corresponding author.}

\begin{abstract}

Multi-agent systems powered by Large Language Models (LLMs) are an emerging paradigm for tackling complex reasoning tasks. In the critical domain of medicine, these systems show immense promise by simulating expert consultations to improve diagnostic accuracy. However, the substantial computational overhead of existing frameworks critically undermines their real-world applicability, creating a major bottleneck for clinical deployment. This inefficiency stems from a ``one-size-fits-all'' approach that applies intensive multi-agent collaboration to all cases, regardless of their intrinsic difficulty. To address this challenge, we propose ConfAgents, an adaptive multi-agent framework. The core insight of ConfAgents is a principled, two-stage process: it first employs a robust, confidence-based triage mechanism to reliably identify and escalate only the most complex cases for collaborative deliberation. For these escalated cases, ConfAgents then initiates an enhanced collaborative process where agents can dynamically retrieve and integrate external knowledge to surmount the limitations of their static training. Extensive experiments on four medical benchmarks demonstrate that ConfAgents not only achieves state-of-the-art accuracy but also drastically reduces computational costs, for instance by over 50\% on the MedQA dataset, moving these powerful systems significantly closer to practical clinical use.

\end{abstract}

\begin{links}
  \link{Code}{https://github.com/PKU-AICare/ConfAgents}
\end{links}

\section{Introduction}

Multi-agent systems driven by Large Language Models (LLMs) represent a leading paradigm for complex reasoning, particularly in the high-stakes domain of medical decision-making (MDM)~\cite{kim2024mdagents}. By simulating the collaborative deliberation of a Multidisciplinary Team (MDT)~\cite{taberna2020multidisciplinary}, these frameworks can synthesize diverse perspectives to achieve a level of diagnostic accuracy and robustness that a single model often cannot~\cite{tang2023medagents, chen2023reconcile}. As the complexity of modern medicine grows with the proliferation of new knowledge~\cite{masic2022medical}, the ability of multi-agent systems to perform intricate, collaborative reasoning establishes them as a vital frontier in medical AI.

However, this enhanced capability comes at a significant, and often prohibitive, cost. Current multi-agent frameworks are universally challenged by a critical efficiency bottleneck that severely undermines their practical applicability in clinical settings. The core of this issue lies in their ``one-size-fits-all'' approach: they apply an intensive, resource-heavy collaborative process to every case, regardless of its intrinsic difficulty. Our analysis reveals that a multi-agent consultation can be up to 50 times slower than a single LLM call, yet the performance gains are often marginal for the majority of cases (Table~\ref{tab:processing_time_comparison}). We find that a large portion of medical questions can be answered with high confidence and accuracy by a single agent. This indiscriminate application of collaboration leads to substantial and unnecessary consumption of computational resources and introduces significant latency, making these systems impractical for routine clinical deployment.

Inspired by triage processes in human clinical practice, where expert consultations are reserved for genuinely challenging cases, we argue for an adaptive strategy. Such an approach selectively engages the full multi-agent framework only for difficult problems, thereby maximizing efficiency without compromising diagnostic accuracy. However, implementing this strategy presents two significant challenges: (1) \textit{Unreliable Confidence Estimation.} A reliable mechanism to differentiate between simple and complex cases is required. LLMs, however, are notoriously miscalibrated~\cite{da2024llm, ye2024benchmarking} and often express high confidence in erroneous conclusions~\cite{yadkori2024believe, feng2024diverseagententropy}. Naively trusting an agent's self-reported confidence is therefore an inadequate foundation for a robust triage system. (2) \textit{Insufficiency of Static Knowledge.} For complex cases that warrant collaboration, the agents' deliberation is fundamentally constrained by their static, pre-trained knowledge~\cite{ng2025rag, wang2024biorag}. When faced with novel or nuanced problems, existing frameworks lack an intrinsic mechanism to recognize the boundaries of their collective knowledge and dynamically seek external, up-to-date evidence~\cite{li2024enhancing, dos2024domain}.

To address these challenges, we propose ConfAgents, an adaptive multi-agent framework for cost-efficient medical diagnosis. ConfAgents introduces a pivotal module, the CP Judger, which leverages conformal prediction (CP) to generate prediction sets with rigorous statistical guarantees~\cite{cherian2024large, zhou2025conformal}. This allows it to act as a reliable and theoretically grounded gatekeeper, escalating only the low-confidence cases for collaborative consultation. For these complex cases, our framework initiates an enhanced collaborative process armed with an iterative dynamic Retrieval-Augmented Generation (RAG) mechanism. This empowers agents to recognize knowledge gaps during deliberation and to dynamically seek, integrate, and reason upon external evidence from a medical corpus. This dual-component design enables ConfAgents to drastically reduce computational overhead by filtering out simpler cases while simultaneously enhancing diagnostic accuracy on complex ones. Our primary contributions are summarized as follows:
\begin{itemize}
    \item \textit{Insightfully,} we are the first to systematically analyze and address the efficiency-accuracy trade-off in medical multi-agent frameworks. We propose a novel adaptive triage mechanism based on conformal prediction, which provides a statistically rigorous criterion for initiating collaboration.
    \item \textit{Methodologically,} we develop a collaborative mechanism that incorporates an iterative dynamic RAG process. This allows agents to deconstruct a problem, perform targeted evidence retrieval from a medical corpus, and engage in informed deliberation, thereby overcoming the limitations of static knowledge.
    \item \textit{Experimentally,} we conduct extensive experiments on four benchmark medical question-answering datasets. The results demonstrate that ConfAgents achieves state-of-the-art diagnostic accuracy while being significantly more efficient, with up to 7.71$\times$ faster processing time compared to existing multi-agent methods.
\end{itemize}

\section{Preliminary and Motivation}

\begin{figure*}[!ht]
  \centering
  \includegraphics[width=1.0\linewidth]{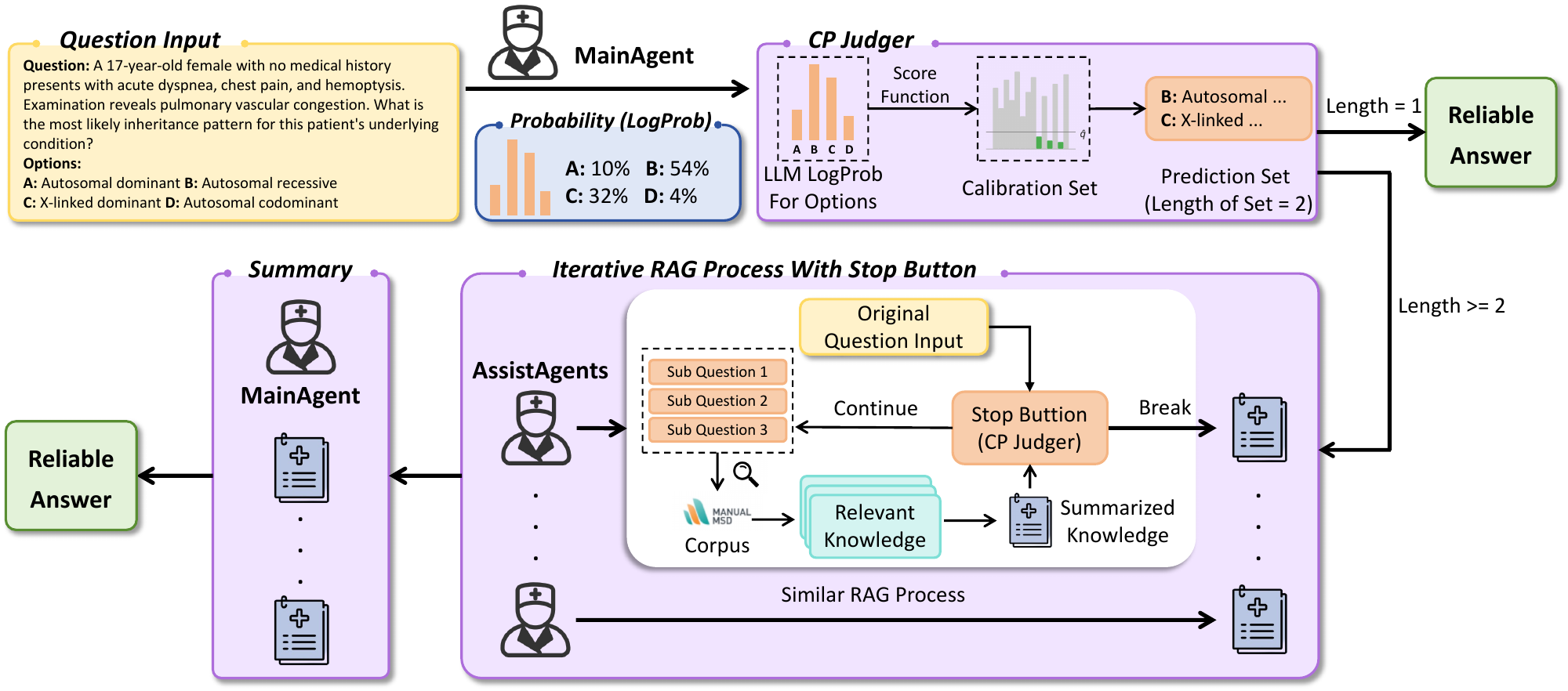}
  \caption{The overall architecture of our proposed ConfAgents framework. It begins with a MainAgent's initial diagnosis. A CP Judger then assesses the confidence; if low, it triggers a collaborative stage where AssistAgents use iterative RAG to gather evidence. Finally, the MainAgent synthesizes all information for a refined diagnosis.}
  \label{fig:model}
\end{figure*}

\subsection{LLM Multi-Agent Collaboration in Medicine}

The application of LLM-driven multi-agent systems, wherein multiple specialized agents collaborate and leverage external tools~\cite{kim2024mdagents}, has emerged as a promising frontier in the medical domain. A significant line of research focuses on creating simulated clinical environments to benchmark agent capabilities, a concept realized in platforms like AI Hospital~\cite{fan2024ai} and Agent Hospital~\cite{li2024agent}. These systems construct virtual hospitals where multiple agents emulate real-world clinical workflows, facilitating the comprehensive evaluation of an LLM's clinical aptitude. Concurrently, another stream of research explores intricate collaboration mechanisms to enhance diagnostic accuracy. Notable examples include MedAgents~\cite{tang2023medagents}, which employs a voting mechanism; ArgMed-Agents~\cite{hong2024argmed}, which resolves conflicts using formal deduction; and ReConcile~\cite{chen2023reconcile}, which introduces a confidence-weighted voting scheme to achieve consensus.

However, while these prior works focus on designing increasingly sophisticated methods to improve performance on medical Q\&A tasks, they often overlook a critical trade-off. We find that such approaches introduce a significant efficiency and resource bottleneck. As illustrated in Table~\ref{tab:processing_time_comparison}, a multi-agent consultation can incur a time overhead nearly 50 times greater than a single LLM generation under a representative setting, while the corresponding performance improvement is often marginal. This disparity highlights a crucial challenge: the substantial benefits of complex consultation are confined to a small fraction of difficult cases, but their high computational cost is paid universally, diluting the overall value and hindering practical adoption.

\begin{table}[!ht]
\centering
\resizebox{\columnwidth}{!}{%
\begin{tabular}{lcccc}
\toprule
\multicolumn{1}{c}{\multirow{2}{*}{\textbf{Methods}}} & \multicolumn{2}{c}{\textbf{Single LLM Agent}} & \multicolumn{2}{c}{\textbf{Multi-LLM Agents}} \\
& Zero-Shot &  Self-Consistency & MedAgents & ColaCare \\
\midrule
Processing Time & 2.1s              & 10.6s      & 113.4s     & 98.3s    \\
Accuracy & 48.6\%              & 51.2\%      & 58.8\%     & 60.8\%    \\
\bottomrule
\end{tabular}
}
\caption{Comparison of processing times and accuracy for single LLM and multi-agent frameworks on the MedQA dataset. The experiment uses GPT-4o and highlights the significant increase in latency for collaborative methods.}
\label{tab:processing_time_comparison}
\end{table}

\subsection{Conformal Prediction}
\label{sec:cp}

Conformal prediction is a statistical framework that provides rigorous, model-agnostic uncertainty quantification for machine learning models~\cite{angelopoulos2021gentle}. A primary advantage is its distribution-free nature, meaning its validity guarantees hold without restrictive assumptions about the underlying data distribution.

\paragraph{Score function.}
The mechanism of CP relies on a non-conformity score function, denoted $g : \mathcal{X} \times \mathcal{Y} \mapsto \mathbb{R}$, which quantifies how ``unusual'' a candidate output $y$ is for a given input $x$. By convention, lower scores signify a better fit. For classification tasks, a standard choice for the score function is one minus the softmax probability of the candidate class, $g(x, y) = 1 - p(y|x)$.

\paragraph{Prediction sets.}
Given a score function $g$ and a threshold $\tau$, the prediction set for a new input $x \in \mathcal{X}$ is formed by including all possible outputs whose non-conformity scores are below this threshold:
\begin{equation}
    C(x; g, \tau) := \{y \in \mathcal{Y} : g(x, y) \leq \tau\}.
    \label{eq:pred_set}
\end{equation}
Intuitively, the size of the prediction set reflects model uncertainty: larger sets imply greater uncertainty, while a singleton set indicates high confidence.

\paragraph{Split conformal prediction.}
We employ the standard split conformal prediction method for its simplicity and computational efficiency. This approach leverages a held-out calibration dataset, $D_{\text{cal}} = \{(x_i, y_i^*)\}_{i=1}^{n_{\text{cal}}}$, to empirically determine the threshold $\tau$. First, we compute the non-conformity scores for the ground-truth pairs in the calibration set:
\begin{equation}
    S_{\text{cal}} = \{ s_i = g(x_i, y_i^*) \mid (x_i, y_i^*) \in D_{\text{cal}} \}.
\end{equation}
For a user-specified miscoverage rate $\alpha \in (0, 1)$, for instance $\alpha=0.1$ for 90\% coverage, the threshold $\tau$ is set to the empirical $(1-\alpha)$-quantile of these scores. With the standard finite-sample correction, it is defined as:
\begin{equation}
    \tau = \operatorname{Quantile} \left(S_{\text{cal}}, \frac{\lceil (n_{\text{cal}}+1)(1-\alpha) \rceil}{n_{\text{cal}}} \right).
    \label{eq:quantile_threshold}
\end{equation}
This calibrated threshold is then used to construct prediction sets for new test points using Equation~\ref{eq:pred_set}, providing a marginal coverage guarantee of at least $1-\alpha$ on unseen data.

\subsection{Uncertainty Quantification for LLMs}

Uncertainty quantification is a foundational research area in machine learning, critical for high-stakes applications like medical decision-making~\cite{da2024llm, ye2024benchmarking}. Early efforts to quantify LLM uncertainty focused on heuristic approaches. A common strategy is to query the LLM for its own confidence score, but models are often poorly calibrated and prone to overconfidence~\cite{yadkori2024believe}. Other approaches leverage metrics like output entropy, but these can be equally unreliable.

Recently, there has been a growing interest in applying conformal prediction (CP) for robust uncertainty quantification~\cite{gligoric2024can, farr2024llm}. Unlike methods that produce a single point estimate, CP generates a prediction set containing all plausible labels. The size of this set serves as a direct and intuitive measure of uncertainty: a larger set implies greater uncertainty. Critically, the resulting set is statistically guaranteed to contain the true label with a user-defined probability, making CP a highly robust and theoretically grounded method.

\section{Methodology}

The proposed LLM-based multi-agent framework, ConfAgents, is illustrated in Figure~\ref{fig:model}. It takes a textual disease description as input and outputs a refined medical diagnosis through a potential four-stage process. Initially, a MainAgent performs a preliminary analysis to generate a probability distribution over potential diagnoses. These probabilities are fed into a CP Judger module, which constructs a conformal prediction set to quantify diagnostic uncertainty. If the prediction set contains more than one option, indicating high uncertainty, the framework triggers a collaborative consultation. In this stage, a team of AssistAgents is engaged to further analyze the case by iteratively interacting with professional medical guidelines to gather additional evidence. Finally, the MainAgent synthesizes its initial assessment with the new evidence from the consultation to produce the final, refined diagnosis.

\subsection{MainAgent's Initial Diagnosis}
Given an input query $x$, which comprises patient information and a set of candidate options $\mathcal{Y} = \{y_1, \dots, y_K\}$, the LLM-based MainAgent, $\mathcal{A}_{\text{main}}$, simulates a medical expert to perform a preliminary analysis. This process yields a probability distribution $\mathbf{p}$ over the candidate options. For white-box models, these probabilities can be directly extracted from the output logits; for black-box models, they can be approximated via sampling or by parsing logit biases if available.
\begin{equation}
    \mathbf{p} = \mathcal{A}_{\text{main}}(x, \mathcal{Y}),
\end{equation}
where $\mathbf{p} = (p_1, \dots, p_K)$ is a vector of probabilities corresponding to each option $y_k \in \mathcal{Y}$. The initial diagnosis $y_{\text{init}}$ is then the option with the highest probability:
\begin{equation}
    y_{\text{init}} = \arg\max_{y_k \in \mathcal{Y}} p_k.
\end{equation}
This initial assessment serves as a baseline for the subsequent confidence evaluation.

\subsection{CP Judger: Deciding Whether to Collaborate}
The CP Judger quantitatively assesses the model's uncertainty to determine if the initial diagnosis is reliable or if it requires multi-agent collaboration. Following the formalism in Section~\ref{sec:cp}, we define the non-conformity score for a given input $x$ and a candidate option $y$ as $g(x, y) = 1 - p(y|x)$, where $p(y|x)$ is the probability from the MainAgent's initial assessment.

To calibrate our uncertainty estimation, we introduce a hyperparameter $\alpha \in (0, 1)$, which represents the permissible marginal error rate. Our goal is to construct prediction sets that contain the true label with a probability of at least $1-\alpha$. To achieve this, we require a calibration dataset, $D_{\text{cal}}$. Instead of using a fixed global set, we dynamically construct $D_{\text{cal}}$ for each test sample by selecting the $N$ most semantically relevant questions from our knowledge base, ensuring the calibration is tailored to the specific query.

The uncertainty threshold, $\tau$, is then calculated as the $(1-\alpha)$-quantile of the non-conformity scores computed on $D_{\text{cal}}$, following Equation~\ref{eq:quantile_threshold}. Finally, for a new test sample $x_{\text{test}}$, we form a prediction set, $C(x_{\text{test}})$, which includes all candidate options whose non-conformity scores are less than or equal to $\tau$:
\begin{equation}
C(x_{\text{test}}) = \{y \in \mathcal{Y} \mid g(x_{\text{test}}, y) \leq \tau\}.
\label{eq:prediction_set_method}
\end{equation}
The size of this prediction set, $|C(x_{\text{test}})|$, serves as our decision criterion for collaboration:
\begin{itemize}
    \item If $|C(x_{\text{test}})| = 1$, it signifies a high-confidence prediction. The MainAgent's initial judgment is accepted and output directly.
    \item If $|C(x_{\text{test}})| > 1$, it indicates high uncertainty, as multiple options are deemed plausible under the specified coverage guarantee.
\end{itemize}
In cases where $|C(x_{\text{test}})| > 1$, the CP Judger flags the sample as uncertain and triggers the collaborative diagnosis protocol.

\subsection{AssistAgents' Collaboration with Iterative RAG}
When the CP Judger identifies a sample as uncertain, the system assembles a team of specialized AssistAgents for an in-depth, collaborative analysis. This process unfolds in two main stages.

\paragraph{Dynamic agent selection.}
The MainAgent first analyzes the candidate options in the prediction set to identify the relevant medical domains (e.g., cardiology, neurology). It then recruits a corresponding specialist AssistAgent for each domain. The set of AssistAgents is dynamically formed as:
\begin{equation}
\mathcal{A}_{\text{Assist}} = \left\{ \operatorname{Agent}(d) \mid d \in \bigcup_{y \in C(x_{\text{test}})} \operatorname{Domain}(y) \right\},
\label{eq:agent_selection}
\end{equation}
where $\operatorname{Domain}(y)$ returns the set of medical domains associated with candidate option $y$.

\paragraph{Iterative RAG for evidence synthesis.}
Each AssistAgent independently executes an iterative RAG loop to build a robust, evidence-based report for its assigned hypothesis. A single iteration for an agent $A_i$ investigating hypothesis $y_i$ involves the following steps:
\begin{enumerate}
    \item \textit{Question Decomposition:} The agent deconstructs its complex diagnostic task into a set of precise sub-questions designed to gather critical evidence.
    \begin{equation}
    \mathcal{Q}_i = \{q_{i,1},  \dots, q_{i,m}\} = \operatorname{SubQuestions}(x, y_i).
    \label{eq:sub_question}
    \end{equation}
    \item \textit{Evidence Retrieval:} For each sub-question $q \in \mathcal{Q}_i$, the retriever component queries the medical corpus, retrieving the top-$k$ most relevant documents $\mathcal{D}_{\text{retrieved}}(q)$ based on semantic similarity.
    \item \textit{Report Synthesis:} The agent filters, extracts, and synthesizes salient information from the retrieved documents $\bigcup_{q \in \mathcal{Q}_i} \mathcal{D}_{\text{retrieved}}(q)$ into a coherent evidence report, $R_i$.
    \begin{equation}
    R_i = \operatorname{Synthesize}(x, \mathcal{Q}_i, \bigcup_{q \in \mathcal{Q}_i} \mathcal{D}_{\text{retrieved}}(q)).
    \label{eq:synthesis}
    \end{equation}
    \item \textit{Iterative Refinement:} The agent formulates a new judgment based on the synthesized report $R_i$. The iterative loop for agent $A_i$ terminates if its judgment becomes unambiguous or a maximum number of iterations is reached. Otherwise, it refines its sub-questions and continues the loop.
\end{enumerate}
Once all AssistAgents complete their iterative processes, their final reports are passed to the aggregation stage.

\subsection{MainAgent's Refined Diagnosis}
In the final stage, the MainAgent aggregates the evidence from all AssistAgents to render a conclusive diagnosis. It receives the set of evidence reports $\{R_1, R_2, \dots, R_m\}$ and the original query $x_{\text{test}}$ as input. Rather than merely tallying the agents' suggestions, the MainAgent performs a holistic synthesis of the arguments and findings within the reports. This evidence-driven decision process is formalized as:
\begin{equation}
y_{\text{final}} = \operatorname{Refine} \left( x_{\text{test}}, \{R_i\}_{A_i \in \mathcal{A}_{\text{Assist}}} \right).
\label{eq:final_diagnosis}
\end{equation}
Here, the $\operatorname{Refine}$ function represents the agent's complex reasoning to produce the single, most probable diagnosis based on a comprehensive review of all collected evidence.

\section{Experimental Setups}

\begin{table*}[!ht]
\small
\centering
\renewcommand{\arraystretch}{1.2}
\begin{tabular}{lcccccccc}
\toprule
\multirow{2}{*}{\textbf{Methods}} & \multicolumn{2}{c}{\textbf{MedQA}} & \multicolumn{2}{c}{\textbf{MMLU}} & \multicolumn{2}{c}{\textbf{MedBullets}} & \multicolumn{2}{c}{\textbf{AfrimedQA}} \\
\cmidrule(lr){2-3} \cmidrule(lr){4-5} \cmidrule(lr){6-7} \cmidrule(lr){8-9}
& ACC ($\uparrow$) & P-Time ($\downarrow$) & ACC ($\uparrow$) & P-Time ($\downarrow$) & ACC ($\uparrow$) & P-Time ($\downarrow$) & ACC ($\uparrow$) & P-Time ($\downarrow$) \\
\midrule
\multicolumn{9}{c}{\cellcolor{gray!15}\textbf{DeepSeek-V3}} \\
\midrule
Zero-Shot & 65.70\stddev{3.55} & 6.66\stddev{1.62} & 76.50\stddev{4.92} & 4.49\stddev{0.19} & 40.40\stddev{4.76} & 4.69\stddev{0.33} & 58.60\stddev{6.77} & 4.04\stddev{0.18} \\
Few-Shot & 63.60\stddev{2.91} & 5.45\stddev{0.34} & 71.80\stddev{4.04} & 4.65\stddev{0.19} & 46.00\stddev{4.31} & 4.90\stddev{0.36} & 58.00\stddev{7.14} & 3.60\stddev{0.11} \\
Self-Consistency & 69.10\stddev{3.14} & 23.63\stddev{1.45} & 75.00\stddev{5.06} & 21.52\stddev{0.67} & 42.50\stddev{4.67} & 24.79\stddev{1.69} & 57.40\stddev{6.58} & 19.47\stddev{0.48} \\
\hdashline
MedAgent & 68.20\stddev{3.34} & 126.05\stddev{4.49} & \textbf{84.40\stddev{3.26}} & 112.82\stddev{6.23} & 57.50\stddev{4.86} & 109.21\stddev{2.34} & 59.70\stddev{6.15} & 96.60\stddev{3.45} \\
ColaCare & 71.40\stddev{3.14} & 140.80\stddev{7.44} & 82.80\stddev{2.36} & 116.06\stddev{6.22} & 54.80\stddev{7.11} & 177.34\stddev{42.24} & 60.70\stddev{4.92} & 118.92\stddev{6.33} \\
MDAgents & 67.30\stddev{4.12} & 94.85\stddev{3.07} & 76.90\stddev{3.24} & \textbf{41.49\stddev{2.26}} & 47.60\stddev{4.10} & 94.95\stddev{1.59} & 61.10\stddev{6.53} & 35.66\stddev{4.21} \\
ReConcile & 66.10\stddev{3.59} & 189.83\stddev{13.36} & 71.60\stddev{3.14} & 153.28\stddev{5.86} & 54.30\stddev{5.71} & 191.69\stddev{6.10} & 53.10\stddev{6.02} & 149.17\stddev{8.39} \\
\hdashline
\textbf{ConfAgents} & \textbf{75.70\stddev{2.90}} & \textbf{65.72\stddev{8.21}} & 80.00\stddev{3.13} & 49.43\stddev{4.11} & \textbf{57.70\stddev{4.49}} & \textbf{43.15\stddev{4.81}} & \textbf{69.50\stddev{6.82}} & \textbf{24.74\stddev{5.83}} \\
\midrule
\multicolumn{9}{c}{\cellcolor{gray!15}\textbf{GPT-4o}} \\
\midrule
Zero-Shot & 56.80\stddev{3.87} & 1.17\stddev{0.05} & 68.00\stddev{4.86} & 1.19\stddev{0.06} & 45.80\stddev{3.68} & 1.24\stddev{0.09} & 47.50\stddev{6.77} & 1.16\stddev{0.07} \\
Few-Shot & 72.40\stddev{3.72} & 1.15\stddev{0.04} & 78.30\stddev{2.53} & 1.05\stddev{0.03} & 68.50\stddev{5.63} & 1.11\stddev{0.08} & 60.90\stddev{6.30} & 1.09\stddev{0.05} \\
Self-Consistency & 49.70\stddev{3.38} & 7.62\stddev{1.84} & 64.30\stddev{3.26} & 5.02\stddev{0.16} & 40.40\stddev{3.07} & 6.23\stddev{1.19} & 45.70\stddev{6.42} & 4.77\stddev{0.11} \\
\hdashline
MedAgent & 78.80\stddev{4.64} & 70.21\stddev{2.56} & 78.80\stddev{3.34} & 71.55\stddev{4.31} & \textbf{72.10\stddev{4.23}} & 73.60\stddev{3.49} & 57.40\stddev{5.75} & 60.92\stddev{4.63} \\
ColaCare & 79.90\stddev{3.83} & 56.74\stddev{4.51} & 79.10\stddev{3.96} & 50.69\stddev{2.74} & 72.00\stddev{4.75} & 54.86\stddev{2.31} & 60.00\stddev{5.66} & 56.50\stddev{2.61} \\
MDAgents & 75.10\stddev{3.99} & 25.15\stddev{2.56} & 80.40\stddev{3.58} & 12.04\stddev{1.41} & 66.70\stddev{6.57} & 32.47\stddev{1.65} & 61.90\stddev{6.12} & \textbf{11.00\stddev{1.73}} \\
ReConcile & 78.10\stddev{2.84} & 46.09\stddev{1.09} & 79.80\stddev{3.43} & 40.70\stddev{0.91} & 70.30\stddev{5.59} & 45.62\stddev{1.02} & 58.90\stddev{5.54} & 42.97\stddev{1.25} \\
\hdashline
\textbf{ConfAgents} & \textbf{80.21\stddev{3.58}} & \textbf{9.11\stddev{2.60}} & \textbf{80.71\stddev{4.01}} & \textbf{6.69\stddev{1.34}} & 70.81\stddev{5.30} & \textbf{30.06\stddev{3.99}} & \textbf{64.58\stddev{5.37}} & 15.31\stddev{3.02} \\
\bottomrule
\end{tabular}
\caption{Overall performance comparison on four medical benchmarks using DeepSeek-V3 and GPT-4o backbones. ConfAgents consistently achieves competitive or state-of-the-art accuracy while being significantly more time-efficient than other multi-agent frameworks, demonstrating a superior balance of performance and cost.}
\label{tab:overall_performance_combined}
\end{table*}

\subsection{Datasets}
We evaluate our model on four multiple-choice medical question answering datasets: (1) MedQA~\cite{jin2021medqa}, a prominent benchmark derived from professional board exams in the United States and China; (2) MMLU~\cite{hendrycks2020measuring}, a massive multitask benchmark, from which we use the medical-related subtasks; (3) MedBullets~\cite{orthotoolkit}, a question bank from an online platform for USMLE Step examinations; and (4) AfriMedQA~\cite{adebayo2024afrimedqa}, which contains questions from South African board exams, focusing on the African medical context. Following MedAgentsBench~\cite{tang2025medagentsbench}, we create specialized Test-Hard and Test-Normal subsets for each dataset, each with 100 samples.

\subsection{Utilized Medical Guideline}
We use the Merck Manual of Diagnosis and Therapy (MSD)~\cite{porter2011merck} as the external knowledge source for our iterative RAG framework, providing agents with access to professional medical guidelines.

\subsection{Evaluation Metrics}
We evaluate all methods on both performance and computational efficiency. For readability, bold indicates the best result in each column of our tables.
\begin{itemize}
    \item \textit{Accuracy (ACC):} The proportion of correct predictions. Higher is better.
    \item \textit{Processing Time (P-Time):} The end-to-end latency in seconds per query. Lower is better.
    \item \textit{Completion Tokens (C-Tokens):} The number of tokens generated by the LLM. Lower is better.
    \item \textit{Total Tokens (T-Tokens):} The sum of input and output tokens. Lower is better.
\end{itemize}
The behavior of our CP Judger is further analyzed in the sensitivity analysis section using standard conformal prediction metrics like coverage rate and average prediction set size.

\subsection{Baseline Methods}
We compare ConfAgents against both single-agent (under three prompting strategies: zero-shot, few-shot, and self-consistency~\cite{wang2022selfconsistency}) and latest advanced multi-agent baselines (MedAgents~\cite{tang2024medagents}, ReConcile~\cite{chen2024reconcile}, MDAgents~\cite{kim2024mdagents}, and ColaCare~\cite{wang2025colacare}).

\subsection{Implementation Details}

\paragraph{Hardware and software configuration.}
All experiments are conducted on a single NVIDIA RTX 3090 GPU. We use Python 3.9, PyTorch 2.3.1~\cite{paszke2019pytorch}, and PyTorch Lightning 2.3.3~\cite{falcon2019lightning}.

\paragraph{Model and hyperparameters.}
For single-agent baselines, few-shot uses two examples, and self-consistency uses three reasoning paths. For multi-agent baselines, we follow their official settings, typically using 3 agents and 3 rounds of iteration. For our ConfAgents, key hyperparameters include the miscoverage rate $\alpha$, the number of retrieval documents $k$, and the maximum number of RAG iterations. Based on empirical tuning, we set $\alpha=0.05$, $k=3$, and the maximum RAG iterations to 3. Performance is reported as mean$\stddev{}$std, calculated by bootstrapping the test set 100 times.

\section{Experimental Results and Analysis}

\subsection{Overall Performance}
As shown in Table~\ref{tab:overall_performance_combined}, our framework demonstrates superior or competitive accuracy compared to all baselines across both DeepSeek-V3 and GPT-4o backbones. Crucially, ConfAgents achieves these strong results with exceptional computational efficiency. For instance, on MedQA with GPT-4o, ConfAgents attains the highest accuracy (80.21\%) while being 2.76$\times$ to 7.71$\times$ faster than other multi-agent methods. These results consistently show that our adaptive approach provides an excellent trade-off between accuracy and computational cost, validating the core design of our framework.

\subsection{Cost Efficiency}

\paragraph{Absolute computational cost.}
As detailed in Table~\ref{tab:agent_consumption_manual_sci}, ConfAgents significantly outperforms other multi-agent methods in computational cost on the MedQA benchmark. Compared to the most time-efficient baseline, MDAgents, ConfAgents is 2.76$\times$ faster in processing time. Furthermore, it is exceptionally token-efficient, consuming 2.01$\times$ fewer completion tokens and 1.22$\times$ fewer total tokens. This demonstrates that the adaptive strategy guided by the CP Judger effectively minimizes unnecessary computations.

\begin{table}[!ht]
\footnotesize
\centering
\begin{tabular}{lccc}
\toprule
\textbf{Method} & \textbf{P-Time (s) $\downarrow$} & \textbf{C-Tokens $\downarrow$} & \textbf{T-Tokens $\downarrow$} \\
\midrule
MedAgent         & 70.21 & $2.07 \times 10^5$ & $7.11 \times 10^5$ \\
ColaCare         & 56.74 & $1.78 \times 10^5$ & $8.57 \times 10^5$ \\
MDAgents         & 25.15 & $0.66 \times 10^5$ & $2.56 \times 10^5$ \\
ReConcile        & 46.09 & $1.45 \times 10^5$ & $4.80 \times 10^5$ \\
\hdashline
\textbf{ConfAgents}        & \textbf{9.11}  & $\boldsymbol{0.33 \times 10^5}$ & $\boldsymbol{2.10 \times 10^5}$ \\
\bottomrule
\end{tabular}
\caption{Efficiency evaluation of different agent-based methods on MedQA using GPT-4o. ConfAgents substantially reduces latency and token consumption.}
\label{tab:agent_consumption_manual_sci}
\end{table}

\paragraph{Cost-accuracy trade-off evaluation.}
To holistically visualize the trade-off between accuracy and cost, we introduce the Balanced Efficiency Score (BES). This intuitive metric is calculated by normalizing both accuracy and processing time to a [0, 1] scale for a given dataset, and then taking their difference: $\text{BES} = \text{ACC}_{\text{norm}} - \text{Time}_{\text{norm}}$. The resulting score ranges from -1 to +1, where a higher score indicates a more favorable efficiency profile. As shown in Table~\ref{tab:bes_results}, ConfAgents achieves the best or near-best score across all datasets, confirming its superior balance of performance and cost.

\begin{table}[htbp]
\footnotesize
\centering
\resizebox{\linewidth}{!}
{\begin{tabular}{lcccc}
\toprule
\textbf{Methods} & \textbf{MedQA} & \textbf{MMLU} & \textbf{MedBullets} & \textbf{AfriMedQA} \\
\midrule
Zero-Shot & 0.23 & \textbf{0.38} & 0.00 & 0.49 \\
Few-Shot & 0.00 & 0.01 & 0.32 & 0.44  \\
Self-Consistency & 0.52 & 0.15 & 0.01 & 0.28  \\
\hdashline
MedAgent & -0.15 & 0.26 & 0.44 & -0.25 \\
ColaCare & -0.27 & 0.13 & -0.14 & -0.29 \\
MDAgents & -0.08 & 0.16 & -0.09 & 0.51 \\
ReConcile & -0.72 & -1.00 & -0.22 & -1.00 \\
\hdashline
\textbf{ConfAgents} & \textbf{0.61} & 0.36 & \textbf{0.47} & \textbf{0.61} \\
\bottomrule
\end{tabular}}
\caption{Cost-performance analysis using the balanced efficiency score (BES) on GPT-4o. ConfAgents achieves the best or near-best score across all datasets, demonstrating superior overall efficiency.}
\label{tab:bes_results}
\end{table}

\subsection{Ablation Study}
We conduct an ablation study to investigate the contribution of our key components on the MedQA dataset. We test three variants: (i) \textit{w/o Iterative RAG}, which uses single-round RAG; (ii) \textit{w/o CP Judger}, which replaces our conformal judger with a naive confidence threshold; and (iii) \textit{w/o Adaptive}, which deactivates the adaptive mechanism and uses the full collaborative process for all problems.

The results in Table~\ref{tab:ablation_study} show that removing the iterative RAG (\textit{w/o Iterative RAG}) hurts accuracy, confirming that iterative evidence gathering is crucial for complex cases. Removing the CP Judger (\textit{w/o CP Judger}) leads to poor decision-making, resulting in a massive increase in processing time as it needlessly triggers collaboration. Finally, removing the adaptive mechanism (\textit{w/o Adaptive}) yields high accuracy but at a prohibitively high computational cost, highlighting the necessity of our triage approach.

\begin{table}[!ht]
\footnotesize
\resizebox{\linewidth}{!}{\begin{tabular}{lcccc}
\toprule
\multicolumn{1}{c}{\multirow{2}{*}{\textbf{Methods}}} & \multicolumn{4}{c}{\textbf{MedQA}} \\
& ACC ($\uparrow$) & P-Time ($\downarrow$) & C-Token ($\downarrow$) & T-Token ($\downarrow$) \\
\midrule
w/o Iterative RAG & 65.50\stddev{3.32} &10.78\stddev{2.45} & 11093 & 74552 \\
w/o CP Judger & 67.80\stddev{4.07} & 109.87\stddev{11.68}& 153545 & 861917\\
w/o Adaptive & 76.50\stddev{3.14} & 193.29\stddev{2.13} & 277204 & 1561235\\
\hdashline
\textbf{ConfAgents} & \textbf{75.70\stddev{2.90}} & \textbf{65.72\stddev{8.21}} & \textbf{97268} & \textbf{556503} \\
\bottomrule
\end{tabular}}
\caption{Ablation study results on MedQA (DeepSeek-V3 backbone). Each component of ConfAgents provides a significant contribution to its overall performance and efficiency.}
\label{tab:ablation_study}
\end{table}

\subsection{Sensitivity Analysis}
We assess the robustness of ConfAgents by evaluating its sensitivity to the choice of the underlying LLM and the hyperparameter $\alpha$.

\paragraph{Sensitivity to different LLMs.}
We benchmark ConfAgents with four distinct LLMs: DeepSeek, GPT-4o, GPT-4o-mini, and Kimi-K2. As shown in Table~\ref{tab:sens_study}, our framework is effective across different base models. Interestingly, with ConfAgents, the more powerful GPT-4o achieves higher accuracy with less processing time than the smaller GPT-4o-mini. This highlights ConfAgents' ability to efficiently leverage powerful models by selectively applying them, making them more cost-effective.

\begin{table}[!ht]
\footnotesize
\resizebox{\linewidth}{!}{\begin{tabular}{lcccc}
\toprule
\multicolumn{1}{c}{\multirow{2}{*}{\textbf{Backbone LLM}}} & \multicolumn{2}{c}{\textbf{MedQA}} & \multicolumn{2}{c}{\textbf{MedBullets}} \\
& ACC ($\uparrow$) & P-Time ($\downarrow$) & ACC ($\uparrow$) & P-Time ($\downarrow$) \\
\midrule
Deepseek-chat & 73.00 & 77.62 & 58.00 & 103.02 \\
GPT-4o & 77.00 & \textbf{9.11} & 66.00 & \textbf{7.02} \\
GPT-4o-mini & 55.00 & 19.01 & 40.00 & 26.68 \\
Kimi-K2 & 66.00 & 116.37 & 52.00 & 144.49 \\
\bottomrule
\end{tabular}}
\caption{Sensitivity analysis of ConfAgents with different backbone LLMs on MedQA and MedBullets. The framework remains effective across various models.}
\label{tab:sens_study}
\end{table}

\begin{figure}[!ht]
\centering
\subfigure[Model accuracy vs. alpha.]{
  \includegraphics[width=0.45\linewidth]{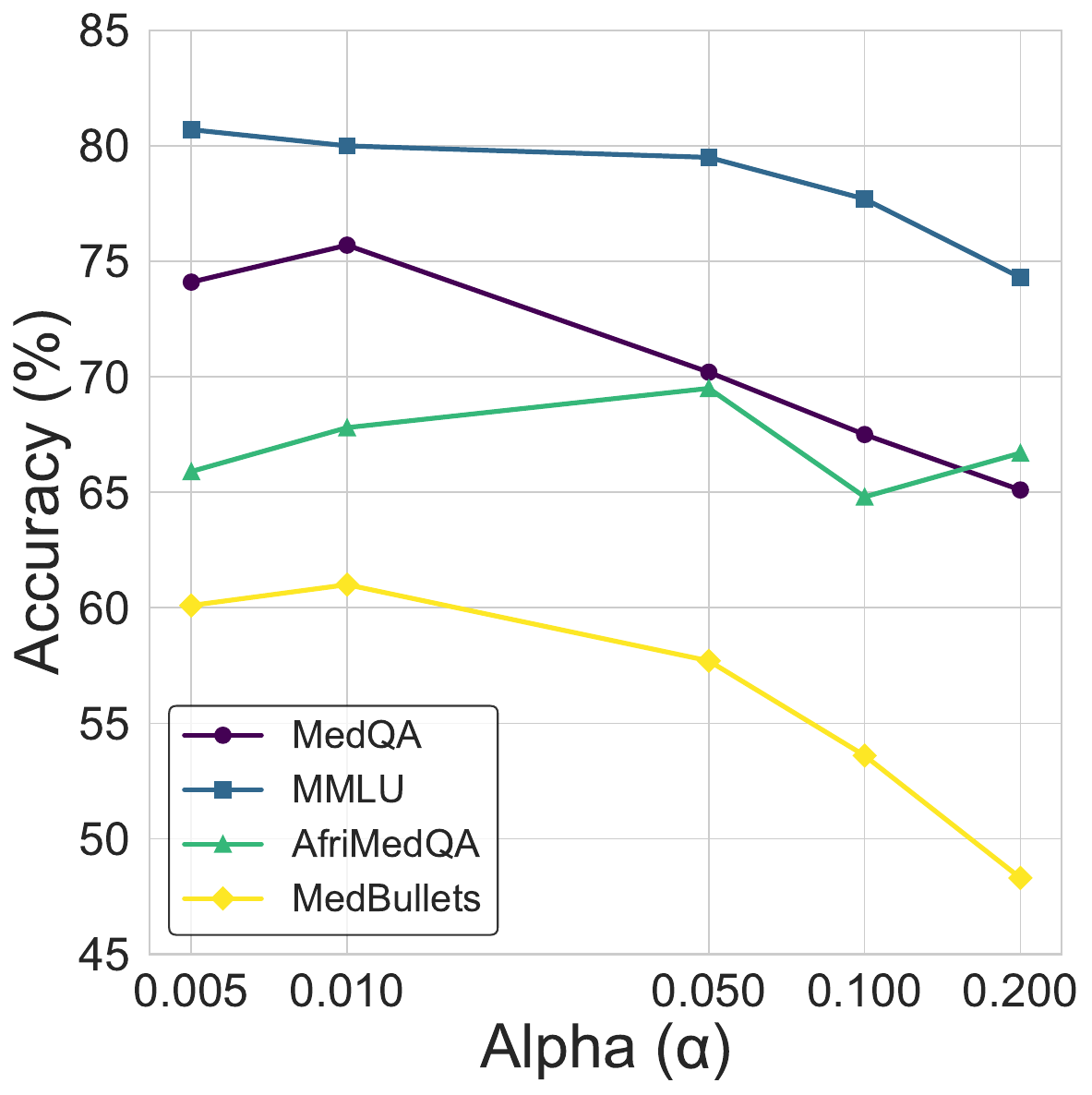}
  \label{fig:acc_vs_alpha}
}
\subfigure[Processing time vs. alpha.]{
  \includegraphics[width=0.45\linewidth]{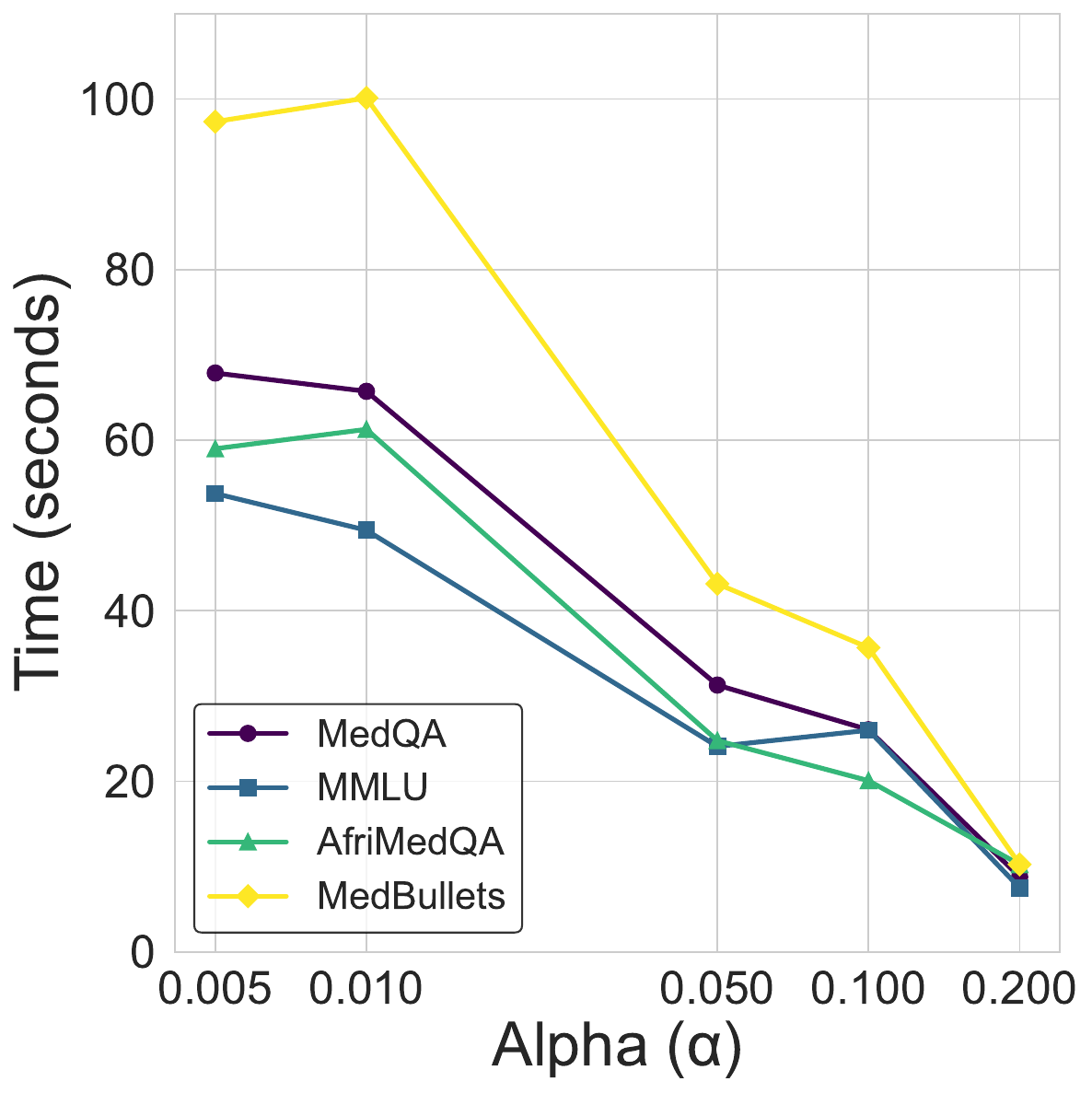}
  \label{fig:time_vs_alpha}
}
\caption{Sensitivity analysis of hyperparameter $\alpha$ on model accuracy (a) and processing time (b) across four datasets. The results show that $\alpha$ provides an effective control over the accuracy-efficiency trade-off.}
\label{fig:acc_time_vs_alpha}
\end{figure}

\paragraph{Sensitivity to hyperparameter alpha.}
The hyperparameter $\alpha$ sets the target miscoverage rate for our CP Judger and directly controls the trade-off between accuracy and cost. A smaller $\alpha$ results in a more cautious system that triggers collaboration more frequently, while a larger $\alpha$ makes the system more aggressive in using the faster, single-agent path. Our analysis, shown in Figure~\ref{fig:acc_time_vs_alpha}, confirms this expected behavior. As $\alpha$ increases, accuracy gradually declines while processing time drops significantly. This demonstrates that $\alpha$ is an effective and interpretable lever to tune the system's behavior for different operational needs. We chose $\alpha=0.05$ for our main experiments as it offers a strong balance between high performance and efficiency.

\subsection{Human Evaluation}

\begin{figure}[!ht]
  \centering
  \includegraphics[width=1.0\linewidth]{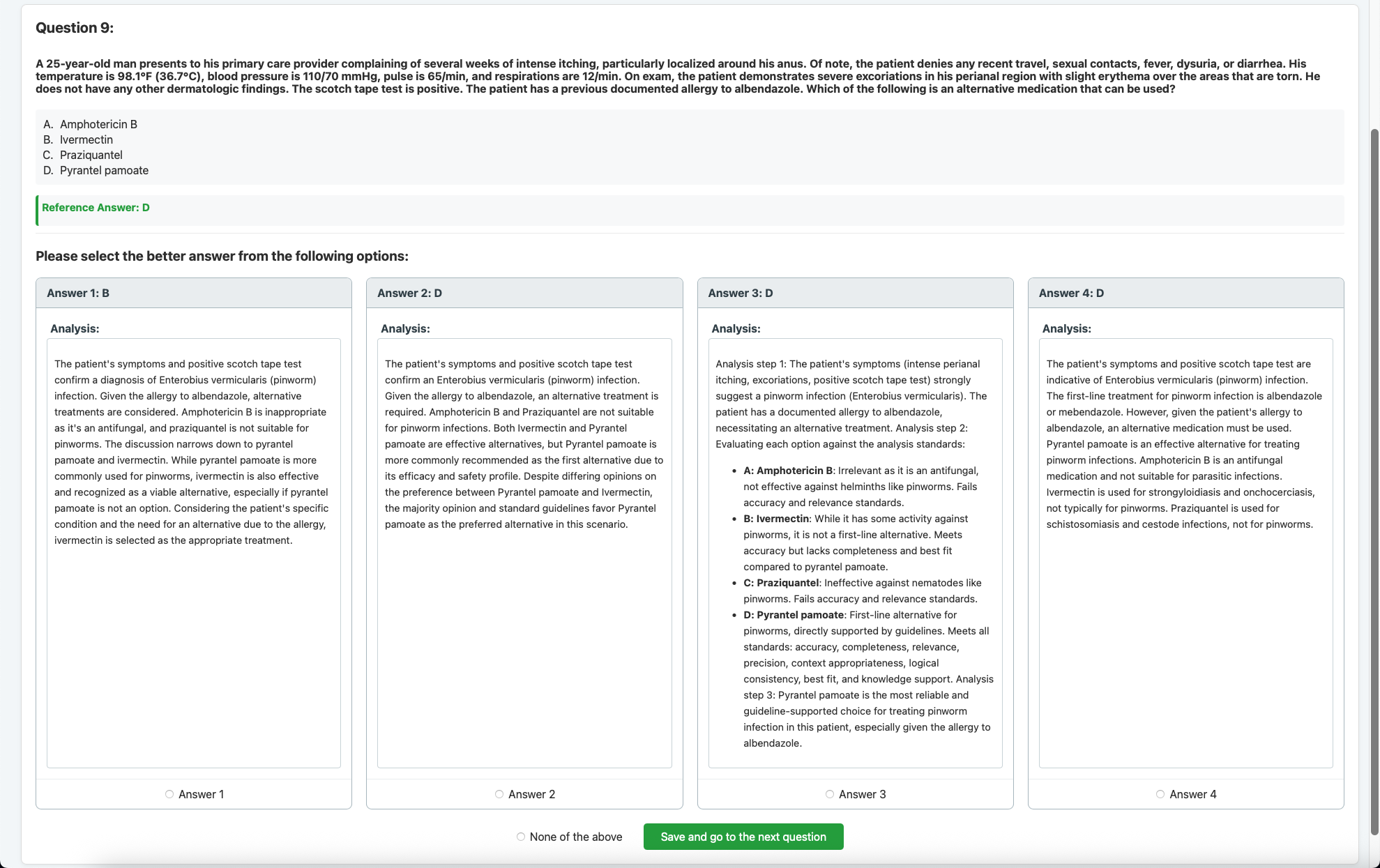}
  \caption{The user interface for our blind human evaluation. Expert evaluators are presented with medical questions and the anonymized, randomly ordered responses from four agent methods (ConfAgents, ColaCare, MedAgent, and MDAgents) to mitigate bias.}
  \label{fig:platform}
\end{figure}

To complement our automated metrics, we conduct a rigorous human evaluation to assess the practical quality of answers generated by ConfAgents. We develop a custom evaluation platform (see Figure~\ref{fig:platform}) and enlist 14 expert evaluators—MD or PhD students in AI for healthcare, bioinformatics, clinical medicine, and biomedical engineering—for a blind, head-to-head comparison on a random subset of 16 questions, 8 from MedQA and 8 from MedBullets.

The results reveal an overwhelming preference for ConfAgents. Our method achieves an average preference rate of 95.54\% across all experts and datasets. The preference is consistent, with a median of 100.0\%, and 9 out of 14 experts (over 64\%) prefer the answers from ConfAgents in all 16 cases. Even the expert with the lowest preference rate still chooses our method for 68.75\% of the questions. This consensus from human experts underscores that ConfAgents not only excels on automated benchmarks but also produces answers that are significantly more coherent, trustworthy, and clinically valuable, reinforcing its potential for real-world application.

\section{Discussion}

\paragraph{Limitations.}
While ConfAgents demonstrates a strong balance of accuracy and efficiency, its practical deployment is contingent upon addressing several key limitations. First, the CP Judger's reliability hinges on a well-matched calibration dataset; for out-of-distribution cases lacking analogous precedents in the knowledge base, the judger's confidence estimates may be unreliable, leading to suboptimal triage. Second, the quality of the agents' deliberation is fundamentally bound by the external knowledge source. The reasoning is only as sound as the retrieved data, and any inaccuracies or gaps in the medical corpus can propagate into the final diagnosis. Finally, our collaboration protocol simplifies the complex, interactive nature of real-world clinical consultation. It lacks mechanisms for direct debate and argumentation among agents, which could be critical for resolving ambiguous cases with conflicting evidence.

\paragraph{Future work.}
These limitations directly inform several promising avenues for future work. We plan to explore more advanced uncertainty quantification, such as class-conditional conformal prediction for finer-grained control and explicit out-of-distribution detection to ensure safety. The collaboration protocol could be significantly enhanced by implementing structured debate frameworks, enabling agents to critique and refine arguments interactively, thereby more closely emulating the dialectic process of expert teams. A crucial next step is developing a human-in-the-loop system, where the CP Judger naturally flags the most uncertain cases for escalation to a human clinician, creating a partnership that optimizes for both efficiency and safety. Finally, the framework's core principle of adaptive, conformal-gated collaboration is highly generalizable, and we will investigate extending this approach to other high-stakes domains such as legal analysis and financial auditing.

\section{Conclusion}

This paper introduces ConfAgents, an adaptive multi-agent framework designed to resolve the critical efficiency-accuracy trade-off in AI-driven medical consultation. By integrating a CP Judger module based on conformal prediction, ConfAgents can reliably determine when to initiate resource-intensive collaboration. For complex cases, its iterative dynamic RAG mechanism empowers agents with external knowledge to enhance their reasoning. Experiments on four medical QA datasets demonstrate that ConfAgents achieves state-of-the-art accuracy while drastically reducing latency and computational overhead. By establishing a new benchmark for efficiency, ConfAgents advances the development of AI decision support systems that are not only powerful but also practical for real-world clinical deployment.

\bibliography{ref}

\appendix

\section{Ethical Statement}

Our research exclusively utilizes publicly available and fully de-identified datasets, such as MedQA, ensuring that no Protected Health Information (PHI) was handled and posing no risk to patient privacy. The ConfAgents framework is developed strictly as a research tool to augment the reasoning process of healthcare professionals and is not intended for direct clinical use, to provide medical advice, or to substitute for professional judgment. Given the nature of the data and the intended research-focused application, we do not foresee any direct ethical risks associated with this work.

\section{Data Availability and Details}

\subsection{Motivation for Dataset Selection}
We select four diverse, multiple-choice question-answering benchmarks for real-world clinical scenarios. Critically, we adopt the sampling strategy from MedAgentsBench~\cite{tang2025medagentsbench}. This decision is driven by its provision of high-quality, curated question sets, with a particular focus on difficult cases. Such challenging problems serve as a better proxy for the complex scenarios where synergistic expert consultation is required, thereby providing a more effective evaluation of our framework's capabilities.

\subsection{Dataset Descriptions and Availability}
For each of the four datasets, we create specialized Test-Hard and Test-Normal subsets, containing 100 samples, to facilitate a nuanced evaluation of performance on problems of varying difficulty.
All datasets employed in this paper are publicly available and are used under their respective data use agreements.
\begin{itemize}
    \item \textbf{MedQA}~\cite{jin2021medqa}: A prominent benchmark featuring questions from professional medical board exams in the United States (USMLE) and China, designed to test high-level medical knowledge.

    \item \textbf{MMLU}~\cite{hendrycks2020measuring}: A massive multitask benchmark intended to measure knowledge acquired during pre-training. We utilize the medical-related subtasks, including anatomy, clinical knowledge, and professional medicine.

    \item \textbf{MedBullets}~\cite{orthotoolkit}: A question bank sourced from an online educational platform for the USMLE Step 1, 2, and 3 examinations, commonly used for medical student training and assessment.

    \item \textbf{AfriMedQA}~\cite{adebayo2024afrimedqa}: A recently developed dataset comprising questions from South African medical board exams, specifically designed to evaluate reasoning within the African medical context.
\end{itemize}

\section{Implementation Details}

\subsection{Computing Infrastructure}

All experiments are conducted on a single Nvidia RTX 3090 GPU with CUDA 12.5. The server's RAM size is 128GB. We implement the model in Python 3.8.18, PyTorch 2.0.0, PyTorch Lightning 2.4.0.
All calls to online LLM APIs (e.g., DeepSeek, Kimi, and GPT) use their respective official platforms. For models that support it, we leverage an OpenAI-compatible API interface for standardized communication.

\subsection{Implementation of Baseline Methods}

For a fair and rigorous comparison, all baseline methods are configured to the best of our ability according to their original papers and public implementations. Unless otherwise specified, all agent frameworks are powered by the DeepSeek-V3 model to normalize for the effect of the backbone LLM and isolate the performance of the agentic architecture itself.

\paragraph{LLM-based single-agent baselines.}
For single-agent baselines, we employ three distinct prompt strategies: zero-shot, few-shot, and self-consistency.
The few-shot approach simulates two sample examples, one positive and one negative, following the approach in previous work\cite{zhu2024benchmarkingllm}. The self-consistency method selects three possible reasoning paths, followed by a consistency evaluation~\cite{wang2022selfconsistency}.

\paragraph{LLM-based multi-agent baselines.}
Additionally, we include four multi-agent collaboration approaches: \textbf{MedAgents}~\cite{tang2024medagents} firstly enhances the model's medical expertise through multidisciplinary collaboration and role-playing. \textbf{ReConcile}~\cite{chen2024reconcile} facilitates multi-round discussions among diverse LLM agents. \textbf{MDAgents}~\cite{kim2024mdagents} classify each medical query into different categories and design different expert consultation modes. \textbf{ColaCare}~\cite{wang2025colacare} integrates multi-agent methodologies targeted at structured electronic health record data.

We primarily use the official repositories for the four multi-agent collaboration approaches, reproducing results according to the instructions provided in their README files. Supplementarily, the input of ColaCare is modified from EHR structured data to textual description in order to fit for our tasks. ReConcile is set to instruct agents to provide confidence scores, and the final prediction is weighted based on these confidence scores.
To ensure a fair comparison, all frameworks are configured to use DeepSeek-V3 as their primary execution model. The number of agents is uniformly set to 3, with 3 rounds of iteration.

\subsection{Evaluation Details}

\paragraph{Overall evaluation and absolute cost efficiency.}
We evaluate all methods based on a suite of metrics assessing both QA performance and computational efficiency. For performance, we use Accuracy. For efficiency, we measure processing time and token consumption, which are critical for deploying LLMs in practice. Lower values are desirable for all efficiency-related metrics.

\begin{itemize}
    \item \textbf{Accuracy (ACC):} This metric measures the proportion of correct predictions for the multiple choice tasks. A higher ACC value indicates superior model performance.

    \item \textbf{Processing Time (P-Time):} This metric quantifies the end-to-end latency, measured from the start of processing a query to the delivery of the final result. It serves as a direct indicator of the method's efficiency.

    \item \textbf{Completion Tokens (C-Token):} This metric counts the number of tokens generated by the model to complete a task. The volume of generated tokens is a primary factor influencing both inference latency and costs.

    \item \textbf{Total Tokens (T-Token):} This metric represents the sum of tokens in the input prompt and the LLM response. It measures the overall computational workload.
\end{itemize}

\paragraph{Cost-accuracy trade-off evaluation.}

To provide a holistic evaluation that considers both predictive accuracy and computational cost, we introduce the \textbf{Balanced Efficiency Score (BES)}. This metric quantifies the trade-off between the performance gain a method achieves and the time cost it incurs, relative to other methods on a given dataset. The score is derived by first normalizing accuracy ($\text{ACC}_{\text{norm}}$) and time ($\text{Time}_{\text{norm}}$) to the $[0, 1]$ scale. $\text{ACC}_{\text{norm}}$ captures the achieved performance gain, while $\text{Time}_{\text{norm}}$ represents the relative time cost. The final BES is calculated as their difference:
\begin{equation}
\label{eq:bes}
\text{BES} = \text{ACC}_{\text{norm}} - \text{Time}_{\text{norm}}
\end{equation}
The resulting score ranges from -1 to +1. A higher BES indicates a more favorable efficiency profile.

\section{Prompt Details}

\subsection{MainAgent Initial Judgment Prompt}
\begin{tcolorbox}[colback=cadmiumgreen!5!white,colframe=forestgreen!75!white,breakable,title=\textit{MainAgent Initial Judgment Prompt.}, label=1]
\begin{VerbatimWrap}
You are an experienced medical expert, proficient in field {{field}}. Now, please answer the following QA questions.

The questions is: {{question}} 
The options are: {{options}} 

Please analyze this question carefully and provide the following two objects:
(1) your only answer option
(2) your confidence in each option. Note: The confidence indicates how likely you think the option is true. Your confidence level, please only include the numerical number in the range of 0-1. The sum of the confidence values for all options should equal 1.   

Make sure the output is in JSON format like this: { "Answer": "C", "conf_each_option": { "A": 0.02, "B": 0.03, "C": 0.85, "D": 0.03, "E": 0.07 } }
\end{VerbatimWrap}
\end{tcolorbox}

\subsection{RAG Decompose Question Prompt}
\begin{tcolorbox}[colback=cadmiumgreen!5!white,colframe=forestgreen!75!white,breakable,title=\textit{RAG Decompose Question Prompt.}, label=2]
\begin{VerbatimWrap}
You are an assistant doctor helper. Your task is to help analyze a multiple-choice question comprehensively.
Here is the question: {{main_question}},Options: {{options}}

Please generate three distinct and critical questions that would help provide a thorough analysis of this question from your perspective as {{field}}. The questions should be probing and aim to uncover key information related to the main question and all options.

Return your response as a JSON object with a single key "questions" which is a list of the three generated questions. The questions are then fed into a retrieval system to retrieve relevant information, so they should be formatted as queries for an embedding model/BM25 index, like search engine queries.
Example format: {"questions": ["question 1", "question 2", "question 3"]
\end{VerbatimWrap}
\end{tcolorbox}

\subsection{RAG Analysis Prompt}
\begin{tcolorbox}[colback=cadmiumgreen!5!white,colframe=forestgreen!75!white,breakable,title=\textit{RAG Analysis Prompt.}, label=3]
\begin{VerbatimWrap}
Please analyze the following multiple-choice question based on the provided retrieved context.
Original Question: {{question}}
Options: {{options}}

Retrieval Query: 
{{retrieve_query}}

Retrieved Context:
{{formatted_context}}

Your task is to provide a comprehensive analysis of the question and all options using the given context from your perspective as field {{field}}. Please be aware that the retrieved information (RAG) may be irrelevant or contain inaccuracies, so you must think critically and evaluate the information's reliability. 
Provide your analysis focusing on:
1. Understanding of the question
2. Evaluation of each option
3. Your reasoning process
Output your analysis as structured paragraphs.
\end{VerbatimWrap}
\end{tcolorbox}

\subsection{AssistAgent Judgment Prompt}
\begin{tcolorbox}[colback=cadmiumgreen!5!white,colframe=forestgreen!75!white,breakable,title=\textit{AssistAgent Judgment Prompt.}, label=4]
\begin{VerbatimWrap}
Please act as a meticulous and objective analysis expert. Your task is to conduct an in-depth evaluation of a multiple-choice question and determine the best answer. Your analysis should be rigorous, evidence-based, and strictly confined to the information presented.

**Question:**
{{question}}

**Options:**
{{options}}

**Retrieved Knowledge Analysis:**
{{rag_analysis}}

**Analysis Standards:**
1. **Accuracy**: Is the option factually correct and scientifically sound?
2. **Completeness**: Does the option fully address what the question is asking?
3. **Relevance**: Is the option directly related to the question's core focus?
4. **Precision**: Is the option specific enough and not overly vague or general?
5. **Context Appropriateness**: Does the option fit the context and scope of the question?
6. **Logical Consistency**: Is the option internally consistent and free of contradictions?
7. **Best Fit**: Among correct options, which one best answers the question?
8. **Knowledge Support**: How well does each option align with the retrieved knowledge analysis?

**Instructions:**
Apply the above standards to evaluate each option systematically, incorporating insights from the retrieved knowledge analysis. The retrieved analysis provides additional context and expert knowledge that should inform your evaluation, but you should also apply critical thinking to assess its relevance and accuracy.

Provide your analysis in the following JSON format:
{ "reasoning": ["Analysis step 1 (incorporating retrieved knowledge where relevant)", "Analysis step 2 (evaluating options against standards)", "Analysis step 3 (final decision rationale)"], "answer": "A/B/C/D" }

\end{VerbatimWrap}
\end{tcolorbox}

\subsection{MainAgent Refined Judgment Prompt}
\begin{tcolorbox}[colback=cadmiumgreen!5!white,colframe=forestgreen!75!white,breakable,title=\textit{MainAgent Refined Judgment Prompt.}, label=5]
\begin{VerbatimWrap}
You are an expert in logical reasoning, critical thinking, and decision analysis. Your task is to analyze a given question along with several candidate answers. Each candidate answer is accompanied by a preliminary analysis and additional detailed reasoning.

**Question:**
{{question}}

**Initial Judgement:**
{{initial_judgement}}

** Assist Information**
{{assist_infos}}

**Instructions:**
Carefully review the question, initial answer, and all supporting analyses. Consider the following factors:

1. **Consensus Among Analyses**: Do most analyses point to the same answer?
2. **Quality of Reasoning**: Which analyses provide the most sound and logical reasoning?
3. **Evidence Strength**: Which option has the strongest supporting evidence across analyses?
4. **Initial Answer Validation**: Does the initial answer align with the analytical evidence?
5. **Contradictions Resolution**: How do you resolve any conflicting viewpoints in the analyses?
6. **Overall Coherence**: Which answer creates the most coherent and complete solution?

**Output Format:**
Provide your final decision in the following JSON format:
{
    "best_answer": "A or B or C or D",
    "confidence_level": "high/medium/low", 
    "reasoning": "detailed explanation of why this answer is the best by fully considering the given analyses"
}
Please ensure your analysis is objective, detailed, and based on evidence and logic.

\end{VerbatimWrap}
\end{tcolorbox}

\end{document}

%% file: packages.tex
\usepackage{times}  
\usepackage{helvet}  
\usepackage{courier}  
\usepackage[hyphens]{url}  
\usepackage{graphicx} 
\usepackage{natbib}  
\usepackage{caption} 
\usepackage{algorithm}
\usepackage{algorithmic}
\usepackage{subfigure}

\usepackage{newfloat}
\usepackage{listings}
\DeclareCaptionStyle{ruled}{labelfont=normalfont,labelsep=colon,strut=off} 
\floatstyle{ruled}
\newfloat{listing}{tb}{lst}{}
\floatname{listing}{Listing}
\usepackage{amsmath}
\usepackage{amssymb}
\usepackage{amsfonts}
\usepackage{booktabs}
\usepackage{siunitx}    
\usepackage{nicefrac}
\usepackage{microtype}
\usepackage{xcolor}
\usepackage{subcaption}
\usepackage{multirow}
\usepackage{makecell}
\usepackage{pifont}
\usepackage{enumitem}
\usepackage{arydshln}

\setlist[enumerate]{leftmargin=*, label=(\arabic*), topsep=0pt}
\setlist[itemize]{leftmargin=*, topsep=0pt}
\usepackage{tikz}
\definecolor{yellowtext}{RGB}{68,132,243}
\definecolor{yellowred}{RGB}{50,167,82}
\definecolor{yellowblue}{RGB}{251,191,5}

\newcommand{\stddev}[1]{\textcolor{gray}{\scalebox{.8}{$\pm$#1}}}
\definecolor{darkblue}{rgb}{0, 0, 0.5}
\definecolor{darkgreen}{rgb}{0.0, 0.42, 0.24}
\definecolor{maroon}{HTML}{A00000}
\definecolor{gray}{rgb}{0.5, 0.5, 0.5}
\definecolor{chocolate}{HTML}{D2691E}
\definecolor{maroon}{HTML}{A00000}
\definecolor{indigo}{HTML}{4B0082}
\definecolor{violet}{HTML}{4B2E83}
\definecolor{lightgreen}{HTML}{E0FBE0} 
\definecolor{lightred}{HTML}{FBE0E0}   
\definecolor{lightblue}{rgb}{0.0, 0.0, 0.5}
\definecolor{cadmiumgreen}{rgb}{0.0, 0.42, 0.24}
\definecolor{forestgreen}{rgb}{0.13, 0.55, 0.13}
\usepackage[most]{tcolorbox}
\usepackage{fvextra}
\DefineVerbatimEnvironment{VerbatimWrap}{Verbatim}{
breaklines=true, 
breakindent=0pt, 
breaksymbol={},
fontsize=\scriptsize,
}

\usepackage[framemethod=tikz]{mdframed}
\mdfdefinestyle{customstyle}{
linecolor=cyan!70,
linewidth=3pt,
innerleftmargin=5pt,
topline=false,
rightline=false,
bottomline=false,
leftline=true,
innerrightmargin=5pt,
innertopmargin=5pt,
innerbottommargin=5pt,
backgroundcolor=cyan!8,
}

\usepackage{longtable}
\usepackage{tabularx}
\usepackage{array}
\newcolumntype{P}[1]{>{\raggedright\arraybackslash}m{#1}}
\usepackage{colortbl}
\definecolor{lightgray}{rgb}{0.9, 0.9, 0.9}

\usepackage{pgfplots}
\pgfplotsset{compat=1.18} 
